\documentclass[]{typhoon} 

\usepackage{amsmath,amssymb}
\usepackage[toc,page,header]{appendix}
\usepackage{minitoc}
\usetikzlibrary{patterns}
\usepackage{float}
\usepackage{bm}
\usepackage{soul}
\usepackage{algorithm}
\usepackage{algpseudocode}
\usepackage{cleveref}
\usepackage{enumitem}
\usepackage{tabularx}
\usepackage{booktabs}
\usepackage{multirow}
\usepackage{graphicx}
\usepackage{xspace}

\newcommand{\huggingface}{\raisebox{-1.5pt}{\includegraphics[height=1.05em]{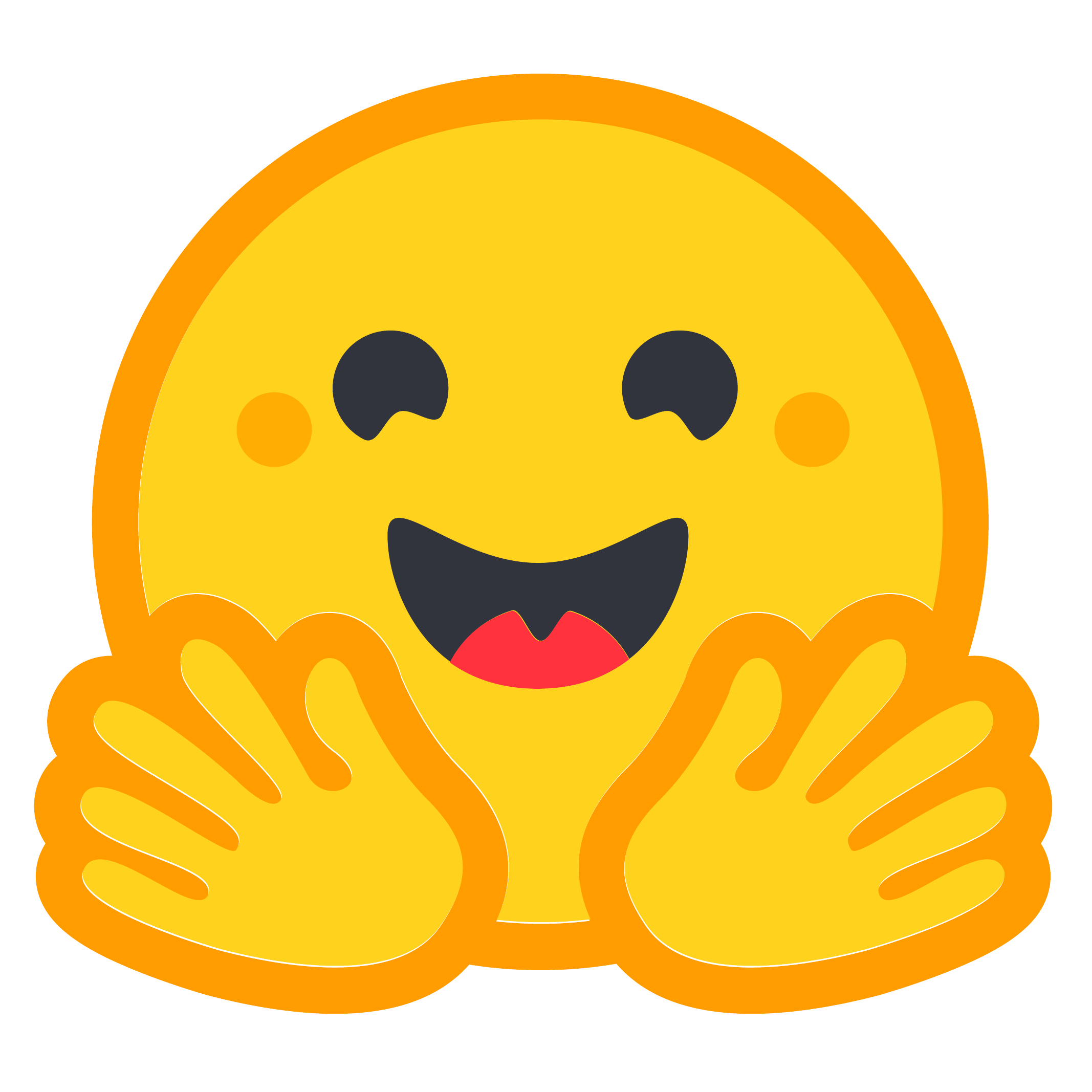}}\xspace}
\newcommand{\github}{\raisebox{-1.5pt}{\includegraphics[height=1.05em]{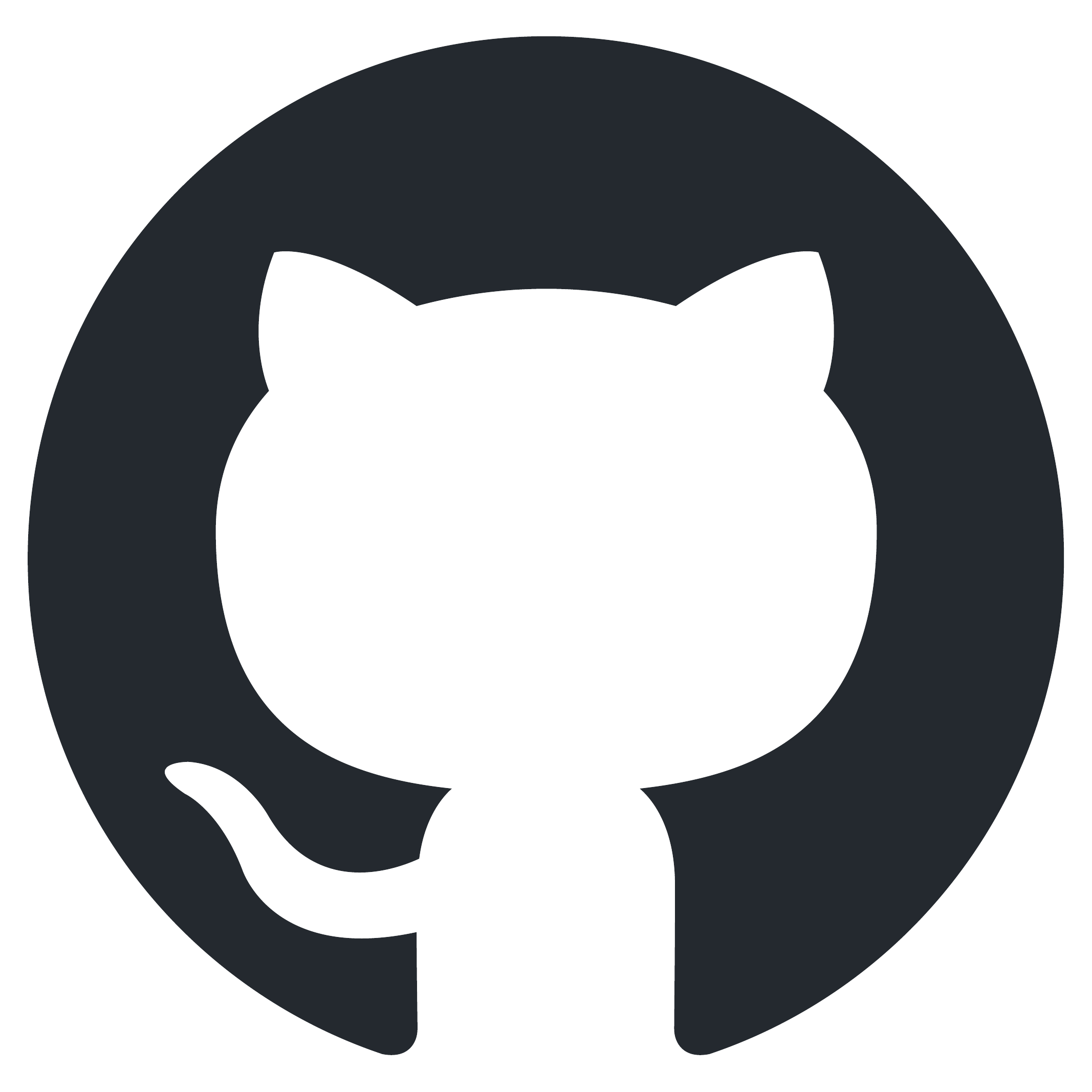}}\xspace}
\newcommand{\typhoon}{\raisebox{-1.5pt}{\includegraphics[height=1.05em]{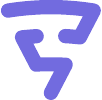}}\xspace}

\setlist[itemize]{leftmargin=*}
\setlist[enumerate]{leftmargin=*}

\title{Typhoon ASR Real-time: FastConformer-Transducer for Thai Automatic Speech Recognition}
\author{Warit Sirichotedumrong, Adisai Na-Thalang, Potsawee Manakul, \\ Pittawat Taveekitworachai, Sittipong Sripaisarnmongkol, Kunat Pipatanakul}
\affiliation{Typhoon, SCB 10X}

\abstract{
Large encoder-decoder models like Whisper achieve strong offline transcription but remain impractical for streaming applications due to high latency. However, due to the accessibility of pre-trained checkpoints, the open Thai ASR landscape remains dominated by these offline architectures, leaving a critical gap in efficient streaming solutions. We present \textbf{Typhoon ASR Real-time}, a 115M-parameter FastConformer-Transducer model for low-latency Thai speech recognition. We demonstrate that rigorous text normalization can match the impact of model scaling: our compact model achieves a \textbf{45$\times$ reduction in computational cost} compared to Whisper Large-v3 while delivering comparable accuracy. Our normalization pipeline resolves systemic ambiguities in Thai transcription--including context-dependent number verbalization and repetition markers (\textthai{ๆ}; \textit{mai yamok})--creating consistent training targets. We further introduce a two-stage curriculum learning approach for Isan (north-eastern) dialect adaptation that preserves Central Thai performance. To address reproducibility challenges in Thai ASR, we release the \textbf{Typhoon ASR Benchmark}, a gold-standard human-labeled datasets with transcriptions following established Thai linguistic conventions, providing standardized evaluation protocols for the research community.
}

\checkdata[Project Resources]{\newline
\github\texttt{Code:} \url{https://github.com/scb-10x/typhoon-asr} \newline
\huggingface\texttt{Typhoon ASR Real-time}: \url{https://huggingface.co/typhoon-ai/typhoon-asr-realtime} \newline\typhoon\texttt{Website}: \url{https://opentyphoon.ai/model/typhoon-asr-realtime}
}

\begin{document}
\maketitle
\vspace{-20pt}

\section{Introduction}
Automatic Speech Recognition (ASR) has experienced transformative progress in recent years, driven by advances in self-supervised learning and large-scale multilingual modeling. Models such as Wav2Vec2 \citep{baevski2020wav2vec} and XLS-R \citep{babu2021xls} demonstrated the power of pre-training on massive unlabeled audio corpora, while OpenAI's Whisper \citep{radford2023robust} established new benchmarks for transcription accuracy across dozens of languages. However, despite these achievements, state-of-the-art ASR systems face critical limitations when deployed in production environments requiring real-time streaming, particularly for morphologically complex and low-resource languages like Thai.

\textbf{Production-Grade ASR Demands Beyond Accuracy.} While encoder-decoder architectures like Whisper excel at offline transcription accuracy, they are fundamentally not well-suited for low-latency streaming applications. The autoregressive decoding process introduces unpredictable latency, and these models are prone to hallucinations--generating plausible but incorrect outputs, particularly for numbers, dates, and domain-specific terminology. For Thai, a tonal language without explicit word boundaries, these issues are compounded by unique linguistic challenges: ambiguous number representations, context-dependent repetition markers, and inconsistent handling of loanwords. These problems are not adequately addressed by model architecture alone; they require systematic data preparation.

\begin{figure}[t]
    \centering
    \includegraphics[width=1.0\linewidth]{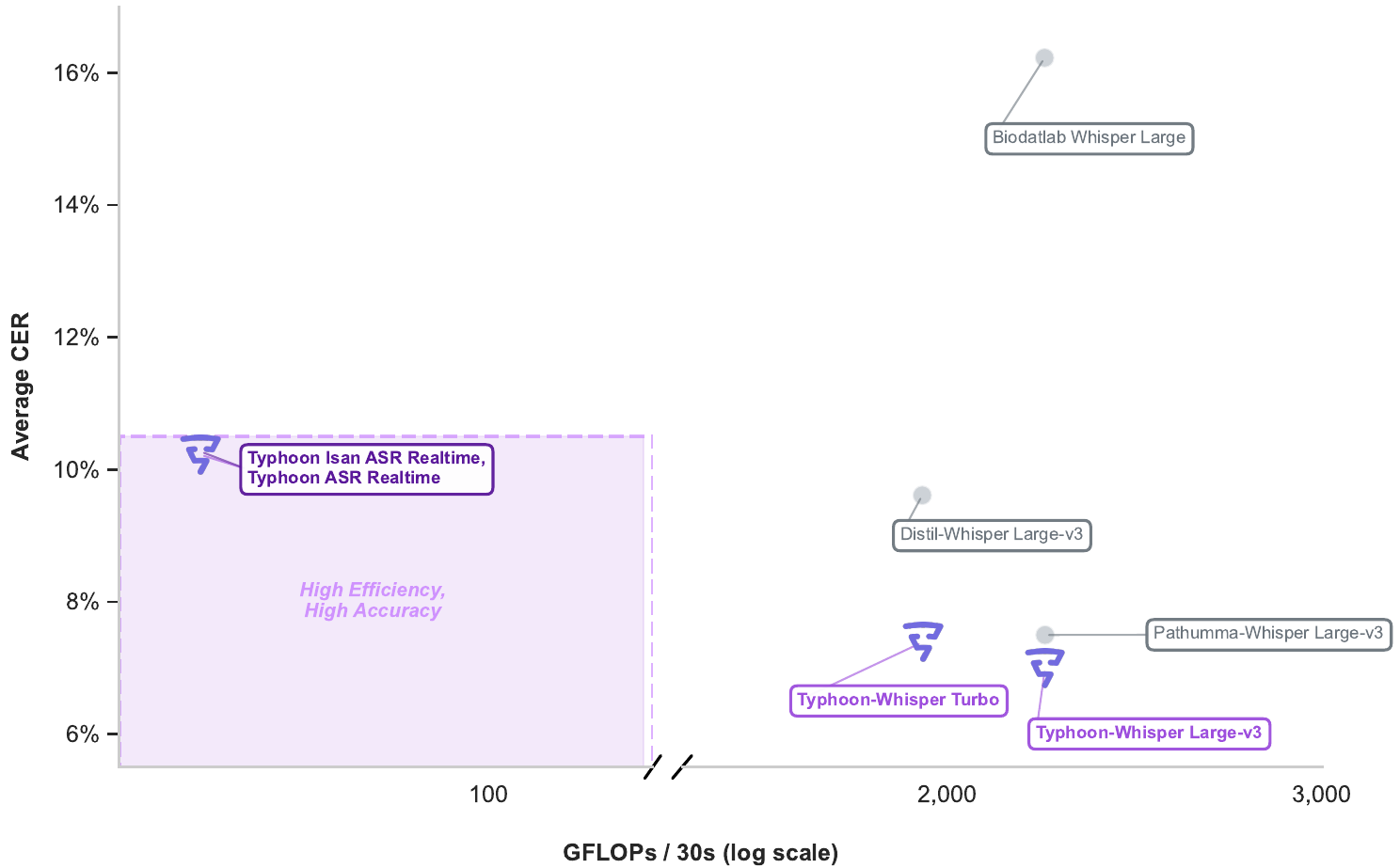}
    
    \caption{\textbf{Pareto efficiency of Thai ASR models.} 
    The scatter plot illustrates the trade-off between computational cost (x-axis, GFLOPs per 30s audio segment on a logarithmic scale) and average Character Error Rate (y-axis) across three benchmarks. 
    Our proposed streaming models, \textit{Typhoon ASR Realtime} and \textit{Typhoon Isan ASR Realtime}, occupy the optimal lower-left region (highlighted), demonstrating comparable accuracy to the offline state-of-the-art \textit{Pathumma-Whisper Large-v3} baseline while achieving an approximate \textbf{45$\times$} reduction in computational complexity.}
    
    \label{fig:pareto_frontier}
\end{figure}

\textbf{Thai ASR Challenge.} Thai presents distinct difficulties for ASR systems. As a low-resource language relative to English or Mandarin, available training data is limited and often inconsistently annotated. The Thai writing system does not use spaces to delimit words, making tokenization non-trivial and rendering standard metrics like Word Error Rate inapplicable. Moreover, the same written form can correspond to multiple spoken realizations: the digits ``10150'' might be read as a postal code (\textthai{หนึ่งศูนย์หนึ่งห้าศูนย์}; \textit{nueng sun nueng ha sun}) or as a quantity (\textthai{หนึ่งหมื่นหนึ่งร้อยห้าสิบ}; \textit{nueng muen nueng roi ha sip}). Without careful normalization, this inherent ambiguity introduces noise into the training process, leading to unreliable model behavior.

\textbf{Our Approach: Data Quality and Consistency as a First-Class Principle.} We argue that for production-ready ASR, particularly in low-resource settings, \textit{data quality and consistency are as important as model architecture}. While existing work focuses primarily on scaling model parameters or training data volume, we demonstrate that systematic data curation--specifically, comprehensive text normalization--is essential for building robust, reliable systems.

In this work, we introduce \textbf{Typhoon ASR Realtime}\footnote{\url{https://huggingface.co/typhoon-ai/typhoon-asr-realtime}}, a FastConformer-Transducer model optimized for streaming Thai speech recognition. While Transformer-based encoder-decoder architectures (e.g., Whisper) currently dominate the Thai ASR landscape, we argue that the \textit{Conformer} family--despite its proven efficiency in streaming scenarios--remains significantly \textbf{under-explored and under-utilized} for Thai.
Our approach revisits this high-potential architecture, combining its superior real-time inference capabilities with a rigorous data preparation pipeline to unlock performance that rivals massive offline models. We make the following contributions:
\begin{itemize}
\item \textbf{Instead of a Whisper-based model, we chose the FastConformer-Transducer}, specifically engineered for low-latency streaming applications with consistent, hallucination-free outputs. 
\item \textbf{A scalable semi-supervised data pipeline} that leverages consensus from multiple teacher models to curate 11,000 hours of training data with \textbf{minimal human intervention}, overcoming the scarcity of large-scale labeled Thai corpora.
\item \textbf{A comprehensive text normalization pipeline} that resolves systemic ambiguities in Thai transcription, creating a canonical training target.
\item \textbf{A multi-stage dialect adaptation strategy} that extends the model to the Isan (north-eastern) dialect while mitigating catastrophic forgetting of Central Thai.
\item \textbf{Typhoon ASR Benchmark and TVSpeech} that address the reproducibility crisis and metric instability in Thai ASR by introducing standardized evaluation protocols. 
\textbf{Gigaspeech2-Typhoon} provides a rigorous, canonically normalized target to resolve the orthographic ambiguities prevalent in public datasets, ensuring metrics reflect phonetic accuracy rather than formatting luck. 
Complementing this, \textbf{TVSpeech} introduces a challenging, manually curated dataset to bridge the gap between academic ``clean speech'' and the \textbf{acoustically complex}, real-world environments encountered in production.
\end{itemize}

\section{Data Curation and Normalization}
A core principle of our models is that the quality and consistency of training data are paramount. We identified that raw transcripts from public corpora are often inadequate for high-fidelity ASR. To address this, we developed a unified data strategy comprising a strict normalization pipeline, a consensus-based transcription system with human verification, and two distinct data mixtures for general and dialect-specific training.

\subsection{Consensus-Based Transcription Pipeline for Pseudo labeling}
While high-quality data are essential, creating labeled data from scratch is challenging, as modern ASR models require transcription at scale. For example, Whisper was trained on more than 500k hours of speech \citep{radford2023robust}, which is not feasible in our setting. Therefore, we must leverage pseudo-labeling to address this limitation.

\begin{figure}[ht]
    \centering
    \includegraphics[width=\linewidth]{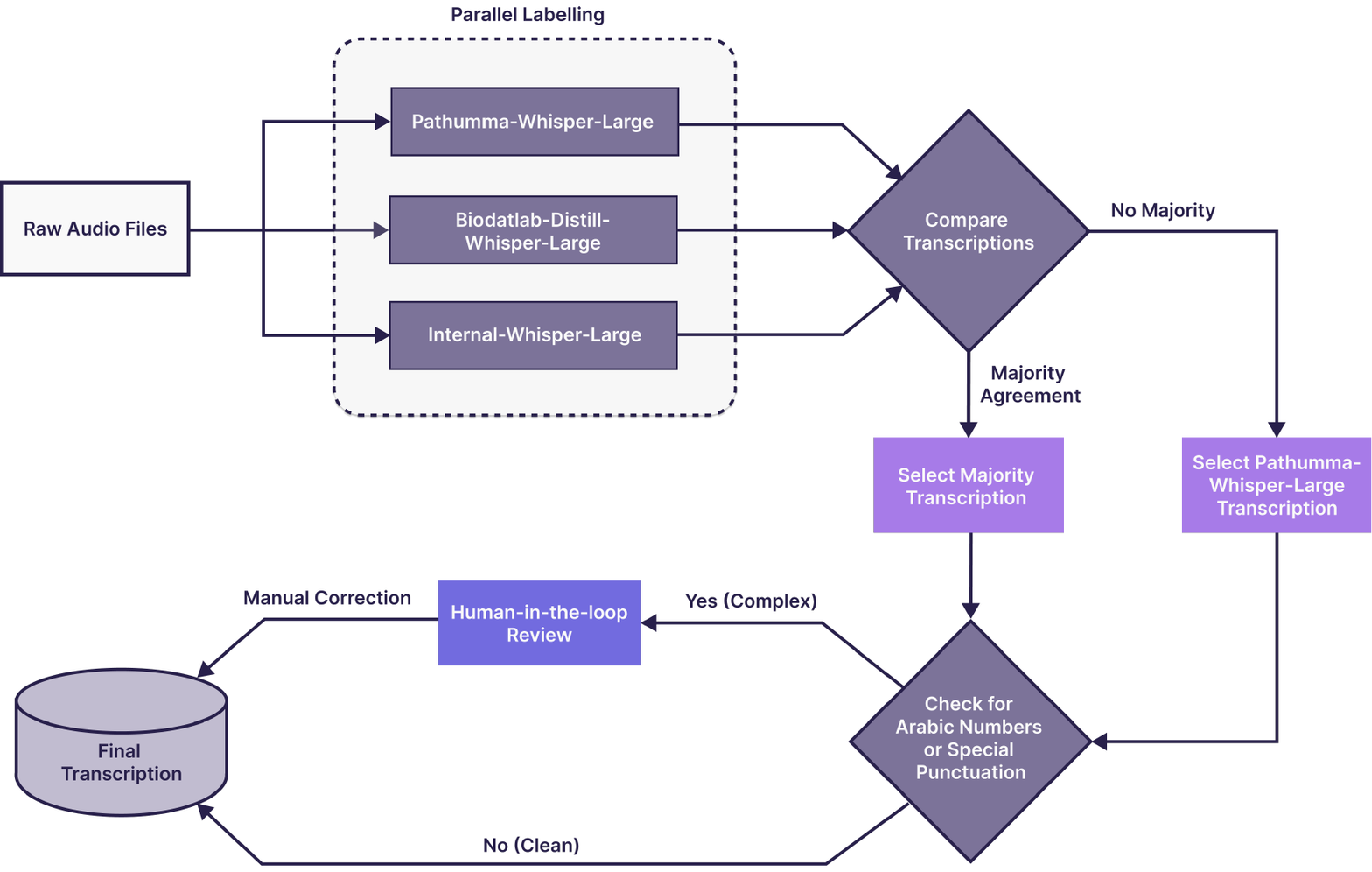} 
    \caption{\textbf{The Consensus Audio Transcription and Verification Pipeline.} Raw audio is processed in parallel by three Whisper-Large models. A majority voting strategy selects consensus transcriptions, defaulting to Pathumma-Whisper-Large when no agreement exists. Complex transcriptions undergo human review, while clean outputs proceed directly to storage.}
    \label{fig:pipeline}
\end{figure}

To ensure high-quality training data at scale, we developed a consensus-based audio transcription and verification pipeline illustrated in \Cref{fig:pipeline}. The pipeline employs a parallel labeling approach where raw audio files are simultaneously processed by three Thai Whisper-Large models: Pathumma-Whisper-Large \citep{tipaksorn2024PathummaWhisper}, Biodatlab-Distill-Whisper-Large \citep{aung-etal-2024-thonburian}, and our Internal-Whisper-Large.

The system implements a majority voting strategy at the comparison stage. When at least two models produce identical transcriptions, that consensus output is selected and proceeds to verification. In cases where no majority agreement exists among the three models, the system defaults to the Pathumma-Whisper-Large transcription as the authoritative output due to its established high performance.

All selected transcriptions undergo an automated complexity verification step that checks for Arabic numerals or special punctuation. Transcriptions flagged as complex are routed to human-in-the-loop review for manual correction, ensuring adherence to our strict normalization rules described in \Cref{sec:text_norm}. Clean transcriptions bypass manual review and proceed directly to the final transcription database. This hybrid approach balances efficiency with quality assurance, maintaining consistency across our training corpus while minimizing human annotation overhead.

\subsection{Text Normalization Pipeline}
\label{sec:text_norm}
Consistent orthography is a prerequisite for robust ASR. We developed a normalization pipeline that transforms text into a canonical representation matching the spoken form. Our pipeline follows the Thai transcription guidelines proposed by \citet{nathalang2025developingopenconversationalspeech}, ensuring linguistic consistency between training targets and evaluation benchmarks. The primary normalization rules and the ambiguities they resolve are detailed in \Cref{tab:normalization_examples}.

\begin{table}[h!]
\centering
\caption{Examples of Text Normalization Rules and Ambiguities}
\label{tab:normalization_examples}
\begin{tabularx}{\textwidth}{@{}llXX@{}}
\toprule
\multirow{2}{*}{\textbf{Rule}} & \multirow{2}{*}{\textbf{Original Text}} & \multirow{2}{*}{\textbf{Ambiguous Interpretation}} & \textbf{Normalized Target}\\ &&&\textbf{(Canonical Form)} \\
\midrule

\multirow{3}{*}{\parbox{2.5cm}{\textbf{Number} \\ \textbf{Standardization}}} 
& `10150' & Spoken as quantity: \textit{nueng muen...} (ten thousand...) & \textthai{หนึ่งศูนย์หนึ่งห้าศูนย์} \newline (\textit{nueng sun nueng ha sun}) \\
\cmidrule(l){2-4}
& \textthai{ไข่เป็ดเบอร์ศูนย์สี่ฟอง} & Written with digit: \textthai{ไข่เป็ดเบอร์ 04 ฟอง} (mixed format) & \textthai{ไข่เป็ดเบอร์ศูนย์สี่ฟอง} \newline (\textit{khai pet boe sun si fong}) \\
\cmidrule(l){2-4}
& \textthai{ตีสามสิบสี่นาที} & Written as \textthai{`ตี} 34 \textthai{นาที'} (implying 3:14 AM) & \textthai{ตีสามสิบสี่นาที} \newline (\textit{ti sam sip si nathi}) \\

\midrule

\multirow{2}{*}{\parbox{2.5cm}{\textbf{Context-Aware} \\ \textbf{Repetition} \\ (\textthai{ๆ}; \textit{mai yamok})}} 
& \textthai{เก่งๆ} & N/A (simple repetition) & \textthai{เก่ง เก่ง} \newline (\textit{keng keng}) \\
\cmidrule(l){2-4}
& \textthai{เป็นอย่างๆ} & Incorrect full-phrase repeat: \textit{pen yang pen yang} & \textthai{เป็น อย่าง อย่าง} \newline (\textit{pen yang yang}) \\

\midrule

\multirow{3}{*}{\parbox{2.5cm}{\textbf{Ambiguous} \\ \textbf{Symbol}}} 
& `6-7' (range) & Read as subtraction: \textit{hok lop chet} & \textthai{หกถึงเจ็ด} \newline (\textit{hok thueng chet}) \\
\cmidrule(l){2-4}
& `6-7' (minus) & Read as a range: \textit{hok thueng chet} & \textthai{หกลบเจ็ด} \newline (\textit{hok lop chet}) \\
\cmidrule(l){2-4}
& `6-7' (separator) & Read as a range: \textit{hok thueng chet} & \textthai{หกขีดเจ็ด} \newline (\textit{hok khit chet}) \\

\midrule

\textbf{Foreign Words} 
& ``website'' & Inconsistent transliterations. & \textthai{เว็บไซต์} \newline (\textit{wep-sai}) \\

\bottomrule
\end{tabularx}
\end{table}

\subsection{General Training Data}
\label{sec:general_data}
Our primary training dataset comprises approximately 11,000 hours of Thai audio. To achieve both broad acoustic coverage and precise output formatting, we constructed a composite dataset that balances massive-scale public corpora with targeted internal datasets.

As detailed in \Cref{tab:data_mixture}, the data mixture is designed to address specific modeling requirements:

\begin{itemize}
    \item \textbf{Foundation (Scale \& Diversity):} We rely on \textbf{Gigaspeech2} as the backbone of our training, providing over 10,000 hours of diverse acoustic environments to learn robust speech features.
    \item \textbf{Robustness \& Benchmarking:} We incorporate an \textbf{Internal Curated Media} dataset--sourced from publicly available content but processed internally--to enhance conversational robustness, alongside \textbf{Common Voice 17.0} \citep{ardila2019common} to ensure performance on standard open benchmarks.
    \item \textbf{Normalization Enforcement:} While large public datasets provide breadth, they often lack consistent text normalization. To address this, we integrated \textbf{Internal TTS (Orpheus)} data specifically generated with complex numeric sequences. This small but critical subset enforces strict adherence to our normalization standards, preventing common hallucinations in number and date transcription.
\end{itemize}

\begin{table}[h]
\centering
\caption{\textbf{General Training Data Composition.} The mixture balances large-scale public corpora with internal curated sets for robustness and precision.}
\label{tab:data_mixture}
\begin{tabular}{l l r r}
\toprule
\textbf{Data Source} & \textbf{Specific Focus} & \textbf{Hours} & \textbf{Utterances} \\
\midrule
Gigaspeech2 & Large-scale Acoustic Diversity & 10,329.5 & 9,843,999 \\
Internal Curated Media & Conversational Robustness & 631.0 & 93,879 \\
Common Voice 17.0 & Read Speech Benchmark & 35.4 & 31,312 \\
Internal TTS (Orpheus) & Numeric Normalization & 3.2 & 2,697 \\
\midrule
\textbf{Total} & & \textbf{10,999.1} & \textbf{9,971,887} \\
\bottomrule
\end{tabular}
\end{table}

\subsection{Isan Dialect Adaptation Data}
\label{sec:isan_data_stats}
For the dialect adaptation stage, we curated a specific dataset of approximately 303 hours. \textbf{Unlike the large-scale general corpus, every utterance in this adaptation set is a gold-standard, human-verified transcription.} We designed this mixture not only to introduce Isan acoustic features but also to maintain robustness in Central Thai and enforce strict output formatting.

As detailed in \Cref{tab:isan_data_mixture_v2}, the data is strategically split between Isan and General Thai sources:

\begin{itemize}
    \item \textbf{Isan Sources (Dialect Signal):} We utilize $\sim$28 hours of our \textbf{Internal Isan Dataset} to capture broad acoustic variety in the dialect, combined with $\sim$28 hours from the \textbf{Public Isan Dataset} (SCB 10X) to cover general and finance-specific domains.
    \item \textbf{General Thai Sources (Regularization \& Formatting):} To prevent catastrophic forgetting and enforce normalization rules, we include $\sim$184 hours of \textbf{Internal Curated Public Media} (General Conversation) for acoustic breadth. We further augment this with $\sim$62 hours of \textbf{Internal TTS} focused on numeric sequences and a targeted \textbf{Public Subset} for repetition markers (\textthai{ๆ}; \textit{mai yamok}). These Central Thai anchors ensure the model retains its ability to handle complex formatting even as it adapts to the new dialect.
\end{itemize}

\begin{table}[h]
\centering
\caption{\textbf{Isan Dialect Adaptation Data Mixture.} A gold-standard mixture of Isan data (for dialect learning) and General Thai data (for regularization and formatting consistency).}
\label{tab:isan_data_mixture_v2}
\resizebox{\linewidth}{!}{
\begin{tabular}{l l l r r}
\toprule
\textbf{Data Source} & \textbf{Language} & \textbf{Specific Focus} & \textbf{Hours} & \textbf{Utterances} \\
\midrule
Internal Curated Media & Central Thai & General Conversation / Acoustic Variety & 184.25 & 25,650 \\
Internal Isan Dataset & Isan & Acoustic Variety & 28.13 & 31,707 \\
Isan Public Dataset & Isan & Recorded General \& Finance Domain & 27.80 & 9,987 \\
Internal TTS & Central Thai & Numeric Formatting & 62.48 & 23,752 \\
Public Subsets & Central Thai & Repetition Markers (\textthai{ๆ}; \textit{mai yamok}) & 0.33 & 106 \\
\midrule
\textbf{Total} & & & \textbf{302.99} & \textbf{91,202} \\
\bottomrule
\end{tabular}
}
\end{table}

\section{Model Architecture and Training Strategy}

\subsection{Model Architecture}
To address the inherent latency and computational bottlenecks of autoregressive models like Whisper \citep{radford2023robust}, we employ the \textbf{FastConformer-Transducer} architecture \citep{rekesh2023fast,nemo_fastconformer_docs}. While Whisper requires processing padded 30-second chunks, our model utilizes a streaming-optimized FastConformer encoder with an 8$\times$ depthwise convolutional subsampling layer (256 channels) and reduced kernel sizes. This aggressive downsampling results in an encoder approximately 2.4$\times$ faster than standard Conformer, enabling efficient frame-synchronous streaming while local attention mechanisms ensure stability on long-form audio.

\subsection{General Thai Model Training}
We fine-tuned all model parameters, initializing from a pre-trained English FastConformer-Transducer (Large) model. Training was conducted for a single epoch on the 11,000-hour General Training Data (\Cref{sec:general_data}) using 2$\times$NVIDIA H100 GPUs with the hyperparameters specified in \Cref{tab:hyperparameters}. The entire fine-tuning process was completed in approximately \textbf{17 hours}, demonstrating the efficiency of the FastConformer architecture even on massive datasets.

\begin{table}[h]
    \centering
    \caption{Training Hyperparameters}
    \label{tab:hyperparameters}
    \begin{tabular}{ll}
    \toprule
    \textbf{Hyperparameter} & \textbf{Value} \\
    \midrule
    Optimizer              & AdamW \\
    Learning Rate Schedule & Cosine annealing (peak 0.001) \\
    Warmup Steps           & 5,000 \\
    Effective Batch Size   & 128 \\
    \bottomrule
    \end{tabular}
\end{table}

\subsection{Curriculum Learning for Dialect Adaptation}
\begin{figure}[htbp]
    \centering
    \includegraphics[width=0.45\linewidth]{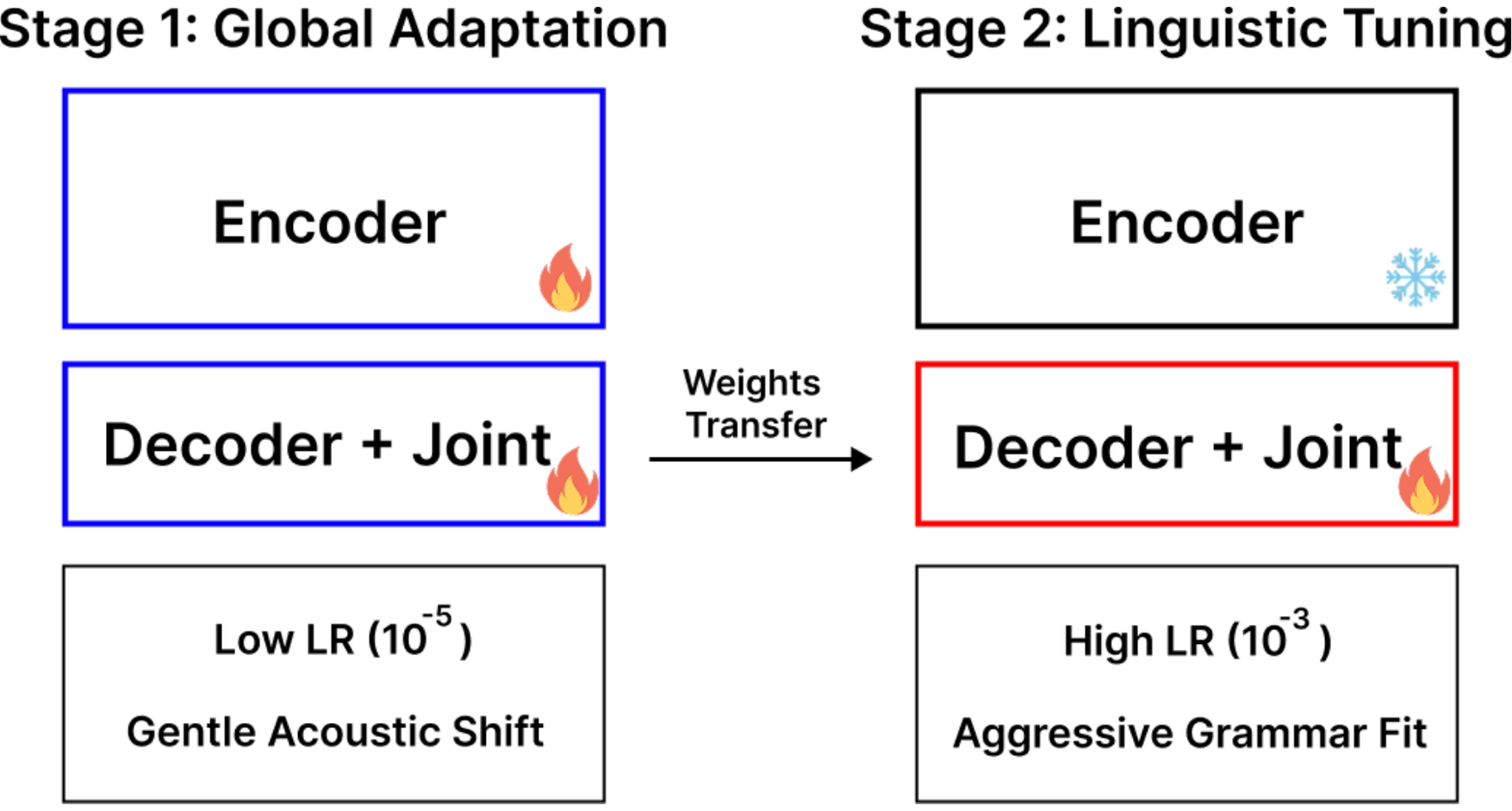}
    \caption{Two-stage curriculum learning for dialect adaptation. Stage 1 employs low learning rate ($10^{-5}$) for gentle acoustic adaptation of the full model over 10 epochs. Stage 2 freezes the encoder (preserving acoustic stability) and employs high learning rate ($10^{-3}$) for rapid linguistic specialization of the decoder and joint network over 15 epochs.}
    \label{fig:multistage_tuning}
\end{figure}

To adapt the model to Isan dialect without catastrophic forgetting of Central Thai, we employ a two-stage curriculum on the 303-hour Isan adaptation dataset (\Cref{sec:isan_data_stats}), as illustrated in \Cref{fig:multistage_tuning}. 

\begin{itemize}
    \item \textbf{Stage 1 (Global Adaptation):} Full-model fine-tuning for 10 epochs with a conservative learning rate ($\eta = 10^{-5}$). This gently adjusts the encoder's acoustic filter banks to capture Isan-specific tonal variations without overwriting robust features learned during Central Thai pre-training.
    
    \item \textbf{Stage 2 (Linguistic Adaptation):} We freeze the encoder and fine-tune only the decoder and joint network for 15 epochs with a higher learning rate ($\eta = 10^{-3}$). This stage focuses on learning Isan lexical structures and dialect-specific particles (e.g., \textthai{บ่} [\textit{bo}; no], \textthai{เฮ็ด} [\textit{het}; do]) while relying on the stable acoustic representations from Stage 1.
\end{itemize}

\section{General Thai Evaluation}
\label{sec:general_eval}
To evaluate the model's performance on standard Central Thai, we utilize the \textbf{Typhoon ASR Benchmark}. This assessment focuses on two critical dimensions: academic accuracy on clean speech and robustness in real-world environments.

\subsection{Typhoon ASR Benchmark}
\label{sec:benchmarks}
To establish a reproducible standard for Thai speech recognition, we introduce \textbf{Typhoon ASR Benchmark}. This benchmark unifies evaluation across two distinct axes: academic strictness and real-world robustness, and both datasets are publicly released for community use. 

Crucially, to ensure deterministic evaluation, all ground truth transcripts in this benchmark have been re-normalized to strictly follow the transcription guidelines proposed in \citep{nathalang2025developingopenconversationalspeech}. This resolves systemic ambiguities (e.g., floating-point pronunciations, repetition markers) that plague existing public leaderboards. Statistics are detailed in \Cref{tab:benchmark_stats}.

\subsubsection*{Remark on Standardization and Fairness}
We acknowledge that applying strict normalization to test sets impacts the absolute scoring of off-the-shelf baselines, which may have been trained on different orthographic conventions (e.g., spelling out numbers vs. using digits). However, we argue that the current lack of standardization in Thai ASR presents a greater threat to evaluation validity. 

Without a canonical target, metrics like character error rate (CER) often penalize models for valid stylistic choices (e.g., ``10'' vs. ``\textthai{สิบ}'' [\textit{sip}]) rather than phonetic errors. By enforcing the standardized guideline from \citet{nathalang2025developingopenconversationalspeech}, we shift the evaluation focus from formatting luck to phonetic accuracy. We encourage future research to adopt this canonical form to ensure consistent, comparable benchmarking across the community.

\subsubsection{Standard Track: Gigaspeech2-Typhoon}
Standard public evaluation on Gigaspeech2 \citep{yang2024gigaspeech2} often suffers from normalization variance. We define the \textit{Standard Track} using a fixed subset of 1,000 utterances (1.01 hours), released as \textbf{Gigaspeech2-Typhoon} \citep{typhoon_gigaspeech_test}. This subset uses our canonical normalization to serve as a stable baseline for clean speech accuracy, ensuring that evaluation focuses on phonetic accuracy rather than formatting inconsistencies.

\begin{table}[!ht]
\centering
\caption{\textbf{The Typhoon ASR Benchmark.} The benchmark comprises two tracks to evaluate both clean-read accuracy and real-world robustness.}
\label{tab:benchmark_stats}
\begin{tabular}{l l c c c}
\toprule
\textbf{Track} & \textbf{Dataset} & \textbf{Hours} & \textbf{Utts} & \textbf{Difficulty} \\
\midrule
Standard & Gigaspeech2-Typhoon \citep{typhoon_gigaspeech_test} & 1.01 & 1,000 & Moderate \\
Robustness & TVSpeech \citep{tvspeech_dataset} & 3.75 & 570 & Hard \\
\bottomrule
\end{tabular}
\end{table}

\subsubsection{Robustness Track: TVSpeech}
To measure performance in ``in-the-wild'' conditions, we release the \textit{Robustness Track} via \textbf{TVSpeech} (Thai Video Speech) \citep{tvspeech_dataset}. This dataset consists of 570 utterances (3.75 hours) curated from diverse public media channels on YouTube under the Creative Commons Attribution (CC-BY) license.

\textbf{Selection Criteria.} Unlike random sampling, this test set was manually curated to maximize both acoustic and semantic complexity. Annotators prioritized segments with high \textbf{lexical density}--specifically selecting clips containing domain-specific terminology, proper names, and technical jargon across categories such as Finance, Technology, and Variety Vlogs. This selection strategy ensures the model is evaluated not just on background noise robustness, but on its ability to accurately resolve low-frequency entities without relying on excessive language model hallucination.

\subsection{Main Results}
\Cref{tab:main_results} compares Typhoon ASR Realtime against offline baselines and foundation models.
\begin{itemize}
    \item \textbf{Competitive Accuracy with High Efficiency:} Typhoon ASR Realtime achieves \textbf{6.81\% CER} on Gigaspeech2-Typhoon, delivering accuracy comparable to the offline Pathumma-Whisper Large-v3 (5.84\%) despite being \textbf{45$\times$ more computationally efficient}. On the \textbf{Thai subset of the} FLEURS benchmark \citep{conneau2023fleurs}, our strict normalization strategy results in higher CER (13.87\%) compared to baselines. This is an expected artifact of orthographic mismatch (e.g., our model outputs ``\textthai{สิบ}'' [\textit{sip}] while FLEURS expects ``10''), rather than phonetic error.
    \item \textbf{Dialect Model Generalization:} The \textit{Typhoon Isan}\footnote{\url{https://huggingface.co/typhoon-ai/typhoon-isan-asr-realtime}} model achieves the best performance across \textit{both} general Thai datasets, demonstrating that our curriculum learning strategy improves general robustness.
    \item \textbf{Foundation Model Variance:} Gemini 3 Pro underperforms on our strictly normalized benchmarks (10.95\% on TVSpeech, 12.50\% on Gigaspeech2). This gap highlights a divergence in objective: while general-purpose foundation models prioritize semantic coherence, they often fail to adhere to the rigid, verbatim transcription standards required for high-precision ASR evaluation.
\end{itemize}

\subsubsection*{Validation of Data Pipeline Superiority}
To isolate the impact of our data curation strategy from architectural differences, we trained two offline baselines on our dataset: \textbf{Typhoon Whisper Large-v3} \citep{typhoon_whisper_large} and \textbf{Typhoon Whisper Turbo} \citep{typhoon_whisper_turbo}. A direct comparison with \textit{Pathumma-Whisper Large-v3}--which shares the exact same model architecture but utilizes a different training corpus--isolates the specific contribution of our data pipeline.

As shown in \Cref{tab:main_results}, our data pipeline yields substantial improvements across both benchmark tracks:
\begin{itemize}
    \item \textbf{Standard Track (Gigaspeech2):} We reduce CER from 5.84\% to \textbf{4.69\%}, representing a relative error reduction of $\sim$20\% on the general academic baseline.
    \item \textbf{Robustness Track (TVSpeech):} The advantage is most pronounced in challenging, in-the-wild environments, where we reduce CER from 10.36\% to \textbf{6.32\%}. This \textbf{4.04\% absolute reduction} confirms that our consensus-based labeling and strict normalization (Section 2) significantly enhance model robustness against background noise and overlapping speech, independent of the underlying neural architecture.
    \item \textbf{The Impact of Normalization on FLEURS:} On the FLEURS benchmark, our models initially show higher CER (e.g., 9.98\% for Typhoon Whisper Large-v3) compared to baselines. However, this is an artifact of orthographic mismatch--our models output spoken forms (e.g., ``\textthai{สิบ}'' [\textit{sip}]) while the benchmark expects digits (``10''). When we evaluate against references normalized to our canonical guideline (shown in parentheses in \Cref{tab:main_results}), our \textit{Typhoon Whisper Large-v3} achieves \textbf{5.69\%} of CER, effectively outperforming both the open-source baselines and the proprietary Gemini 3 Pro (6.91\%). This confirms that the perceived gap is stylistic, not phonetic.
\end{itemize}

\begin{table*}[t]
\centering
\caption{\textbf{Impact of Data Quality on Model Performance.} Comparisons show two distinct advantages: 1) \textbf{Architecture:} Our Streaming model competes with offline baselines. 2) \textbf{Data Pipeline:} When training the exact same architecture (Whisper Large-v3), our data pipeline improves performance by over \textbf{4\% absolute CER} on noisy data (TVSpeech) compared to the state-of-the-art Pathumma baseline. For Fleurs, values in parentheses denote CER evaluated against references normalized via our guidelines, highlighting that high baselines often stem from formatting mismatches rather than phonetic errors.}
\label{tab:main_results}
\setlength{\tabcolsep}{10pt}
\renewcommand{\arraystretch}{1.1}
\resizebox{\linewidth}{!}{
\begin{tabular}{l l c c c }
\toprule
\multirow{2}{*}{\textbf{Type}} & \multirow{2}{*}{\textbf{Model}} & \multicolumn{3}{c}{\textbf{CER (\%)} $\downarrow$} \\
\cmidrule(lr){3-5}

 & & \multirow{2}{*}{\textbf{TVSpeech}} & \multirow{2}{*}{\textbf{Gigaspeech2}} & \textbf{Fleurs} \\
  & & & & \textbf{Orig. (Norm.)} \\
\midrule
\rowcolor{typhoonpurple!20}\multicolumn{5}{l}{\textit{\textbf{Proprietary Foundation Models}}} \\
Offline & Gemini 3 Pro & 10.95 & 12.50 &  11.35 (\underline{6.91})\\
\midrule
\rowcolor{typhoonpurple!20}\multicolumn{5}{l}{\textit{\textbf{Open-Source Offline Baselines}}} \\
Offline & Biodatlab Whisper Large & 18.96 & 13.22 & 16.50 (15.26) \\
Offline & Biodatlab Distil-Whisper Large & 13.82 & 8.24 & \underline{6.77} (8.63) \\
Offline & Pathumma-Whisper Large-v3 & 10.36 & 5.84 & \textbf{6.29} (7.88) \\
\midrule
\rowcolor{typhoonpurple!20}\multicolumn{5}{l}{\textit{\textbf{Ours (Typhoon Data Pipeline)}}} \\
\textbf{Streaming} & \textbf{Typhoon ASR Realtime} & 9.99 & 6.81 & 13.87 (9.68) \\
\textbf{Streaming} & \textbf{Typhoon Isan ASR Realtime} & 9.34 & 6.93 & 14.55 (10.15) \\
\addlinespace
Offline & Typhoon Whisper Turbo & \underline{6.85} & \underline{4.79} & 10.52 (7.08) \\
Offline & Typhoon Whisper Large-v3 & \textbf{6.32} & \textbf{4.69} & 9.98 (\textbf{5.69}) \\
\bottomrule
\end{tabular}
}
\end{table*}
\subsection{Architectural Efficiency and Streaming Capability}
A critical advantage of Typhoon ASR Realtime is its structural suitability for live applications. While offline models maximize accuracy via bidirectional context, our system prioritizes latency and throughput.

\textbf{FastConformer-Transducer.} We adopt the FastConformer-Transducer architecture \citep{rekesh2023fast}, an industry standard for efficient streaming ASR. By utilizing aggressive downsampling in the encoder front-end, this architecture significantly reduces the sequence length passed to attention layers, lowering computational cost (FLOPs) compared to standard Conformers \citep{gulati2020conformer}. Furthermore, our use of a Transducer (RNN-T) decoder \citep{graves2012sequence} enables true streaming inference, avoiding the latency bottlenecks inherent in the autoregressive decoding of Whisper-based models.

\textbf{Parameter Efficiency.} Typhoon ASR Realtime contains only \textbf{115 million parameters}. In contrast, the state-of-the-art offline baseline, Pathumma-Whisper Large-v3, utilizes approximately \textbf{1.55 billion parameters}. Our model is roughly \textbf{13$\times$ smaller}, offering a massive reduction in memory footprint suitable for resource-constrained deployment.

\section{Isan Dialect Evaluation}
To rigorously assess the model's capabilities on the Isan dialect, we conducted both quantitative benchmarking (CER) and qualitative human evaluation (A/B testing).

\subsection{Quantitative Results (Isan Benchmark)}
We evaluated performance on a held-out Isan test set derived from the SCB 10X Thai Dialect Isan Dataset \citep{scb10x_isan}. \Cref{tab:isan_results} compares Typhoon models against external baselines and Gemini 2.5 Pro. To ensure a fair comparison against available open resources, we included \textit{Whisper-Medium-Dialect}, a baseline we trained specifically on the combined public dialect corpora from SLSCU \citep{suwanbandit23_interspeech} and LOTUS-TRD \citep{lotus_trd_2024}.

\begin{table}[h]
\centering
\caption{\textbf{Isan Dialect Character Error Rate (CER).} Offline models generally outperform streaming architectures. The ablation shows that Stage 2 (Linguistic Specialization) is critical, yielding a 5.57\% improvement over Stage 1.}
\label{tab:isan_results}
\begin{tabular}{l l c}
\toprule
\textbf{Model Type} & \textbf{Model Name} & \textbf{CER (\%)} \\
\midrule
Foundation & Gemini 2.5 Pro & 10.20 \\
\midrule
\multirow{3}{*}{Offline} 
 & Whisper-Medium-Dialect & 17.72 \\
 & SLSCU Korat Model \citep{suwanbandit23_interspeech} & 70.08 \\
 & \textbf{Typhoon-Whisper-Medium-Isan} \citep{typhoon_whisper_isan} & \textbf{8.85} \\
\midrule
\multirow{2}{*}{Streaming} & Typhoon Isan ASR Realtime (Stage 1: Acoustic) & 16.22 \\
 & \textbf{Typhoon Isan ASR Realtime (Stage 2: Final)} & \underline{10.65} \\
\bottomrule
\end{tabular}
\end{table}

\subsubsection*{Results Analysis:}
\begin{itemize}
    \item \textbf{Offline vs. Streaming Gap:} The offline \textit{Typhoon-Whisper-Medium-Isan} achieves the lowest CER (8.85\%), benefiting from the full-context attention mechanism.
    \item \textbf{Competitive Streaming Performance:} Our \textit{Typhoon Isan ASR Realtime} achieves 10.65\% CER. While slightly higher than the offline counterpart, it significantly outperforms the \textit{Whisper-Medium-Dialect} baseline (17.72\%)--which was trained on the standard public datasets \citep{suwanbandit23_interspeech, lotus_trd_2024}--demonstrating the superiority of our curriculum learning strategy.
\end{itemize}

\subsubsection*{Ablation Analysis: Impact of Curriculum Learning}
The step-wise improvement in \Cref{tab:isan_results} validates our two-stage adaptation strategy (Section 3.3). 
\begin{itemize}
    \item \textbf{Stage 1 (Acoustic Adaptation):} Adapting the encoder alone yields a CER of 16.22\%. While this captures the tonal distinctiveness of Isan, the model lacks the dialect-specific vocabulary required for accurate transcription.
    \item \textbf{Stage 2 (Linguistic Specialization):} Freezing the encoder and aggressively fine-tuning the decoder/joint network provides a massive \textbf{5.57\% absolute reduction} in CER (16.22\% $\to$ 10.65\%). This confirms that for dialect adaptation, adjusting the language modeling components is as critical as acoustic alignment.
\end{itemize}

\subsection{Qualitative Human Evaluation}
Given the high morphological variance in Isan, CER does not always correlate perfectly with usability. We further validated these results via human A/B testing.

\subsubsection{Protocol and Agreement}
We compared the same systems against \textbf{Gemini 2.5 Pro} (reference) on 500 Isan audio samples. Two native Isan speakers performed blind pairwise comparisons (2,000 total judgments) with moderate inter-annotator agreement (Cohen's Kappa = 0.56).

\subsubsection{Comparative Results}

\begin{figure}[htbp]
    \centering
    \includegraphics[width=\linewidth]{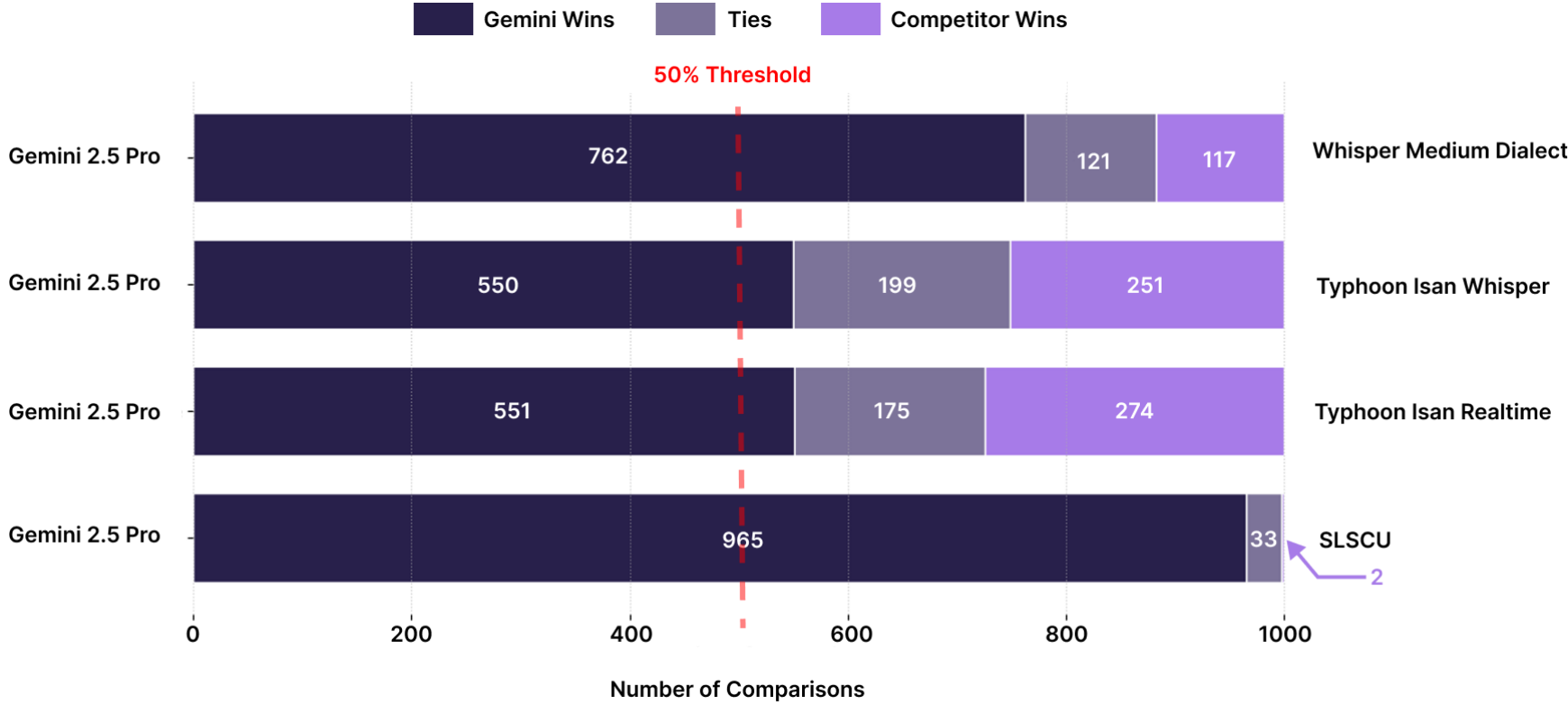} 
    \caption{A/B testing results showing Win-Tie-Loss counts for Gemini 2.5 Pro against various competitor systems (N=1000 comparisons). The dashed red line marks the 500-count (50\%) threshold.}
    \label{fig:gemini_ab_comparison}
\end{figure}

Results (\Cref{fig:gemini_ab_comparison}) show:
\begin{itemize}
    \item \textbf{Gemini Dominance:} Gemini consistently crosses the 50\% win threshold against all competitors, likely due to massive pre-training allowing it to infer semantics even when acoustic signals are ambiguous.
    \item \textbf{Typhoon Superiority among ASR:} Among dedicated ASR systems, our Typhoon Isan models secure the highest combined number of wins and ties.
    \item \textbf{Baseline Failure:} The external SLSCU baseline registers negligible wins, aligning with its high CER (70.08\%) observed in \Cref{tab:isan_results}.
\end{itemize}

\subsection{Analysis: Divergence Between Metrics and Human Preference}
Our results highlight a divergence between automated metrics and human preference. While Gemini 2.5 Pro and Typhoon Isan Realtime have nearly identical CER (10.20\% vs 10.65\%), Gemini wins more often in human preference. This suggests that Foundation Models optimize for semantic coherence (readability), whereas ASR models optimize for phonetic fidelity. However, for a 115M parameter streaming model to perform within 0.45\% CER of a massive Foundation Model demonstrates the effectiveness of our dialect adaptation curriculum.

\section{Limitations and Future Work}

While Typhoon ASR Realtime demonstrates that compact, data-centric models can outperform large-scale offline baselines, we acknowledge specific limitations inherent to our design choices and scope.

\subsection{Limitations}

\textbf{Orthographic Rigidity vs. Readability.} Our strict normalization pipeline (\Cref{sec:text_norm}) prioritizes phonetic fidelity over stylistic readability. For example, the model is trained to transcribe English terms into Thai transliteration (e.g., \textthai{เว็บไซต์} [\textit{wep-sai}] instead of ``website'') and numbers into spoken words (e.g., \textthai{หนึ่งศูนย์} [\textit{nueng sun}] instead of ``10''). While this eliminates ambiguity for the acoustic model, the raw output may not be immediately suitable for end-user display without post-processing.

\textbf{Limited Code-Switching Support.} The current model is optimized for Thai-dominant speech. In scenarios involving heavy Thai-English code-switching--common in technical and corporate environments--the model forces phonetic mapping to Thai characters rather than switching to Latin script. This limits its immediate utility in bilingual domains where mixed-language orthography is preferred.

\textbf{The Semantic Gap.} As observed in the Isan evaluation (Section 5.3), our 115M-parameter model relies primarily on acoustic cues. Unlike massive foundation models (e.g., Gemini), it lacks the deep ``world knowledge'' required to resolve complex semantic ambiguities or context-dependent homophones (\textthai{คำพ้องเสียง}; \textit{kham phong siang}) when the acoustic signal is unclear.

\subsection{Future Directions}

\textbf{Inverse Text Normalization (ITN).} To bridge the gap between phonetic transcription and user-facing applications, future work should develop robust ITN systems that convert spoken-form Thai (e.g., \textthai{หนึ่งศูนย์หนึ่งห้าศูนย์} [\textit{nueng sun nueng ha sun}]) back into written formats (e.g., postal codes, dates, currency) based on contextual cues.

\textbf{Contextual Biasing for Domain Adaptation.} Real-world applications often require recognition of domain-specific vocabulary (e.g., proper names, technical terms, organizational entities) that are underrepresented in training data. Exploring shallow fusion or contextual biasing techniques could enable runtime adaptation without full model retraining.

\textbf{Handling Multi-Speaker Scenarios.} Production deployments require integration with speaker diarization systems to attribute transcriptions in multi-party conversations. Additionally, addressing overlapping speech remains an open challenge for the Thai ASR community.

\textbf{Thai Dialect Coverage.} The success of our Isan adaptation curriculum (Section 5) suggests a scalable path for extending coverage to other regional variants, including Northern (\textit{Kam Mueang}) and Southern (\textit{Pak Tai}) dialects. Investigating unified models capable of zero-shot dialect identification and adaptation would democratize access to voice technology across Thailand.

\textbf{On-Device Deployment.} Given our model's parameter efficiency (115M), exploring quantization (INT8/INT4) and on-device optimization (ONNX, CoreML) could enable privacy-preserving, offline Thai ASR on mobile and IoT devices.

\section{Conclusion}
We introduced \textbf{Typhoon ASR Realtime}, a streaming ASR system that challenges the trend of parameter scaling by prioritizing data quality and architectural efficiency. With only 115M parameters, our model outperforms offline baselines 13$\times$ its size on standard benchmarks, validating our hypothesis that rigorous normalization is essential for low-resource languages.

Our investigation into Isan dialect adaptation revealed a critical divergence between automated metrics and human preference: while Foundation Models like Gemini excel at semantic coherence (high human preference), they struggle with the verbatim precision required for low-CER transcription. Typhoon ASR bridges this gap, offering the phonetic fidelity required for technical ASR tasks with the speed necessary for real-time deployment. We release the \textbf{Typhoon ASR Benchmark}, gold-standard human-labeled dataset with transcriptions following established Thai linguistic conventions, providing standardized evaluation protocols for the research community.

\section*{Acknowledgments}
Beyond the primary authors, we gratefully acknowledge the Typhoon Team members at SCB 10X whose contributions made this project possible: Chanakan Wittayasakpan, Kritsadha Phatcharoen, Sirinya Chaiophat, Surapon Nonesung, Natapong Nitarach, Tanawin Samutsin, Shah Faisal Wani, Krisanapong Jirayoot, Oravee Smithiphol, Kasima Tharnpipitchai, and Kaweewut Temphuwapat. We also extend our appreciation to the SCBx R\&D Team for their support, resources, and valuable insights. Lastly, we are grateful to the global and local AI communities for open-sourcing resources and sharing knowledge.

\bibliographystyle{plainnat}
\bibliography{refs}

\end{document}